\documentclass[12pt]{article}
\usepackage{pdflscape}
\usepackage{multirow}
\usepackage{mathtools}
\usepackage{amsmath}
\usepackage{cite}
\usepackage{amsfonts}
\usepackage{amssymb}
\usepackage{graphicx} 
\usepackage{cite}
\usepackage{multirow}
\usepackage{url}
\usepackage{geometry}
\usepackage{algorithm}
\usepackage{algpseudocode}
\usepackage{comment}
\usepackage{todonotes}

\DeclarePairedDelimiter\floor{\lfloor}{\rfloor}
\begin{document}
\title{K-Bit-Swap: A New Operator For Real-Coded Evolutionary Algorithms}
\author{Aram Ter-Sarkisov\thanks{School of Computing, Dublin Institute of Technology, Ireland, {aram.tersarkisov@dit.ie}}   and 
Stephen Marsland\thanks{School of Engineering and Advanced Technology, Massey University, New Zealand, {s.r.marsland@massey.ac.nz}}}
\maketitle
\begin{abstract}
There has been a variety of crossover operators proposed for Real-Coded Genetic Algorithms (RCGAs), which recombine values from the same location in pairs of strings. 
In this article we present a recombination operator for RCGAs that selects the locations randomly in both parents, and compare it to mainstream crossover operators in a set of experiments on a range of standard multidimensional optimization problems and a clustering problem. We present two variants of the operator, either selecting both bits uniformly at random in the strings, or sampling the second bit from a normal distribution centered at the selected location in the first string. While the operator is biased towards exploitation of fitness space, the random selection of the second bit for swapping makes it slightly less exploitation-biased. Extensive statistical analysis using a non-parametric test shows the advantage of the new recombination operators on a range of test functions.     
\end{abstract}
\section{Introduction}
Genetic Algorithms (GAs) are population-based metaheuristic algorithms. They were first introduced in \cite{holland1975adaptation}, and have demonstrated effectiveness on a wide range of problems, from constrained optimization to design and optimization of neural networks and other classifiers, see e.g. \cite{roeva2013population, pencheva2013genetic}. They are known for their flexibility in terms of the choice of parameters (population and pool sizes, selection function, elitism selection, etc) and encodings: binary, integer and real (float).\\
\linebreak
In the GA framework solutions to a problem are encoded as strings, the fitness of the strings evaluated according to a problem-specific function, and then strings are chosen to enter a `mating pool' as some function of their fitness. Strings in the mating pool are recombined using genetic operators, and these new offspring strings specify a new population. Over many iterations of the algorithm strings with better fitness (that is, better solutions to the problem) evolve. In general, there are two types of genetic operators that are used: crossover and mutation. Mutation varies the values of entries of a single string, and is effectively a version of local search, exploiting the knowledge encoded in the current string and seeking local improvements in it. In contrast, crossover operators recombine genetic information between parents, which enables a form of global search, exploring the search space more effectively. It has been shown both experimentally and (for some problems) theoretically that GAs with versions of crossover and mutation outperform mutation-only algorithms such as hill-climbers and Randomized Local Search (see e.g. \cite{doerr2008crossover, spears1992crossover}).\\
\linebreak
While the original GA was based on a binary string, variations based on integer values and floating point numbers are common. Real-Coded Genetic Algorithms (RCGAs) use chromosomes (strings) of floating point numbers and  are particularly useful for solving problems where `real-valued' encoding arises naturally, e.g. training of classifiers, such as neural networks, see \cite{blanco2001real}, signal processing, constrained optimization problems, etc. They are also particularly useful for higher-dimensional optimization problems, where binary encoding is simply infeasible. Readers are referred to \cite{eshelman1993chapter} for a concise explanation of advantages of RCGAs, and \cite{wright1990genetic} for an extended one.\\
\linebreak
The development and study of new genetic operators for RCGAs is both a historically rich and active topic of research. Some of the seminal papers include \cite{eshelman1993chapter, agrawal1994simulated, deb2002computationally, srinivas1994adaptive, wright1990genetic, michalewicz1991handling}. Recently, in \cite{herrera2003taxonomy, herrera2005hybrid} hybrid crossover operators were thoroughly analyzed, in \cite{yoon2013geometricity} distance-based crossover operators were studied, in \cite{thakur2013new} a double Pareto crossover was introduced and successfully tested on a range of multimodal test functions. Also in \cite{thakur2013new, yoon2012roles} a substantial overview of the history of RCGA crossover operators is given. In many cases parents can generate more (or less) than the standard set of two offspring or an offspring can have more than two parents (see e.g. \cite{ling2007improved, herrera2002multiple, elsayed2011ga}).\\
\linebreak
In this paper we follow in these footsteps, and introduce two real-coded versions of the K-Bit Swap (KBS) genetic operator, which is a crossover operator that enables the location of elements of the string to change (transpose) when crossover occurs. We evaluate its use with a RCGA solving multidimensional multimodal real-coded optimization problems, both alone and together with other genetic operators. A version of KBS was introduced in \cite{tersarkisov2010} (see also \cite{tersarkisov2012thesis} for the extended version of this work) and showed some promise mostly for binary-coded functions. 
\section{Connection with GA Theory, Exploration vs Exploitation at Population and Gene Levels}
\label{sec:theory}
Just as with binary GAs, empirical results show a statistically significant advantage of RCGAs with crossover/recombination operators, when compared with mutation-only algorithms. A few probable reasons for this are that crossovers:
\begin{enumerate}
\item reduce the probability of premature convergence to local optima,
\item extend the list of possible offspring that can be generated (especially when probability distributions over offspring are used),
\item extend the exploration of fitness space to subsets that are far from the current solution,
\item combine exploration and exploitation to some degree, while mutation is primarily an exploitation operator,
\end{enumerate}       
The last point is of particular interest when it comes to development of new crossover operators. In the GA community there is little clarity about the definition of the terms `exploration' and `exploitation'. Intuitively, exploration deals with sampling from entirely new segments of genotype, while exploitation focuses on sampling in the vicinity of existing solutions. In \cite{vcrepinvsek2013exploration} authors give an overview of approaches to this issue since the early years of GA theory. In short, they can be described as follows:   
\begin{enumerate}
\item Genotype-based, e.g. a distance between individuals of some sort (Manhattan, Hamming, etc),
\item Phenotype-based, e.g. a number of different phenotypes (fitness values) in a population that is used to determine diversity.    
\end{enumerate}
Hence, the definitions of exploration and exploitation depend on the definition of diversity in the population: finding new individuals outside of the `basin of similarity' between existing members of the population would be considered exploration. Balancing these two processes, referred to as diversity maintenance (which can be achieved by e.g. niching, crowding, mating restrictions, selection pressure) is crucial to the success of the algorithm. Too much diversity (exploration) can harm by reducing the attention to promising basins around local optima and lack of it (exploitation) reduces the chances of finding good new local optima.\\
\linebreak
Approaches mentioned above are population-level tools. In this paper we try a gene-level approach (as in, e.g. \cite{herrera2002multiple, elsayed2011ga}): given that we have two parents producing two new offspring, there is some distance between the parents and the offspring. Once we have selected the genes/bits for recombination, we use the crossover operator to exchange the information between these genes.  Distance from the parents depends on this operator: if it does search strictly in the landscape between the gene values, it is an exploitation operator; if, instead, the operator extends the search space (but doesn't fully exclude the interval between the gene values), it is biased towards exploration.\\     
\linebreak 
Arithmetical crossover (AX) is the example of the first type of operator (see \cite{michalewicz2013genetic}):
\begin{align*}
h^1_{i} &= \lambda c^{1}_i +(1-\lambda)c^2_{i} \\
h^2_{i} &= \lambda c^{2}_i +(1-\lambda)c^1_{i} 
\end{align*}
where $c^{j}_{i}$ are parental gene values, $0< \lambda<1, \ h^{j}_{i}$ are offspring' gene values. Clearly both $h^{1}_i$ and $h^{2}_i$ lie strictly between $c^{j}_{i}$, so this operator exploits the interval between the sampled values in each parent. A good example of an exploration-biased operator is BLX-$\alpha$ (see \cite{eshelman1993chapter}): 
\begin{align*}
C_{\max} = \max \{c^1_{i}, c^2_{i} \}\\
C_{\min} = \min \{c^1_{i}, c^2_{i} \}\\
I=C_{max}-C_{min}\\
h^{j}_{i} \in [C_{min}-I \alpha, C_{\max} + I \alpha] 
\end{align*}
The last expression means that new values for the $i^{th}$ gene are sampled from this interval, which stretches beyond the interval by a factor of $\alpha$. For example, very high values of this parameter $\approx 1$ almost triple the interval between gene values. In \cite{eshelman1993chapter} this operator was introduced as an example of interval schemata.\\
\linebreak
In Section \ref{sec:kbs} we introduce two new KBS operators that fit the definition of exploitation-biased operators. Nevertheless, we argue that due to their main property they allow a certain degree of landscape exploration.\\
\section{K-Bit-Swap Genetic Operator for Real-Coded Problems}
\label{sec:kbs}
The original K-Bit-Swap operator was developed for predominantly binary-valued GAs. It was presented in \cite{tersarkisov2010}, and a pseudo-code description is given in Figure \ref{fig:tab1}. The operator is a form of crossover where, instead of bits being swapped between the strings with their location in the string held fixed, which is standard with most crossovers, the location of the bit in the new string was chosen uniformly at random. The results in \cite{tersarkisov2010} demonstrated that the operator improved the results of a GA on numerical optimization problems, but not on an integer-coded variant of the Travelling Salesman Problem.
\begin{center}
\begin{figure*}
\begin{algorithmic}
\For{i=1: \textnormal{number of pairs in the pool}}
	\State take the next two strings from the pool
        \For{j=1:K} 
	       \State select a bit in the first string uniformly at random
               \State select a bit in the second string uniformly at random
               \State swap the values of these bits between the two strings
        \EndFor
\EndFor
\end{algorithmic}
\caption{The K-Bit-Swap Operator for binary coding}
\label{fig:tab1}
\end{figure*}
\end{center}
It is perfectly possible to transfer this version of the KBS to real-valued GAs without substantial changes, just like a standard crossover can be transferred. However, given the success of Arithmetical Crossover (AX) and Blend Crossover (BLX-$\alpha$) in RCGAs, we use this convention and consider instead variants of KBS that follow the idea behind these two operators, as well as the original KBS, as described next.
\subsection{$\alpha$KBS}
The main idea behind KBS is to enable swapping of genetic information between different schemata without binding the choice of the location in the second parent to the location selected in the first parent. Discussion of schemata theory in the context of GAs is beyond the scope of this article, relevant information can be found in \cite{goldberg89genetic, mitchell1992royal, mitchell96introduction}. It is only worth mentioning here that unlike the bulk of crossovers, KBS operators recombine information from unrelated schemata. Simple tweaking of the original KBS idea with an additional feature (parameter $\alpha$) similar to the one used in AX and BLX-$\alpha$ is presented in Figure \ref{fig:tab2} (note that the version of KBS described previously is recovered with $\alpha=0$).
\begin{figure*}
\begin{algorithmic}
\State Set $0 < \alpha  \leq 1$ value
\For{i=1:\textnormal{number of pairs in the pool}}
	\State take the next two strings from the pool
        \For{j=1:K}
	       \State select a value in the first string uniformly at random, $v_1$
       	       \State select a value in the second string uniformly at random, $v_2$
               \State set $v'_1 = \alpha v_1 + (1-\alpha)v_2$
               \State set $v'_2 = (1-\alpha)v_1 + \alpha v_2$
	       \State replace the values at these locations: $v_1 = v'_1, \ v_2 = v'_2$
        \EndFor
\EndFor
\end{algorithmic}
\caption{The $\alpha$KBS Operator}
\label{fig:tab2}
\end{figure*}
This operator resembles both AX and BLX-$\alpha$ in the sense that a value in a parent's gene is multiplied by a coefficient ($\alpha$ or $1- \alpha$). However, $\alpha$KBS samples the value to swap with the first parent's value uniformly in the second parent, rather than taking it from the matching location in the string, so that genes from unrelated features (schemata) are swapped. In the current setting, this operator has a greater exploitation, but lesser exploration capacity because the region available for sampling a new value lies strictly between two parental values. This drawback is partly overcome by the main property of randomized bit selection, a feature that AX does not have. Also, unlike BLX-$\alpha$ we do not use an interval around parents' values.    
\subsection{$\beta$KBS}
Next, we consider a variant of KBS that samples from a normal (Gaussian) distribution centered on the chosen bit in the first string to select the bit in the second parent. This `de-randomizes' KBS, making it more of a uniform-type crossover than a mutation operator as, due to lower variance, the selected bit in the second parent is likely to be closer to the original bit. Note that if this normal distribution has 0 variance then this becomes a standard crossover. Empirically, we found the variance of 4 to be a good choice, which is used in the experimental part of the article. 
\begin{figure*}
\begin{algorithmic}
\For{i=1:\textnormal{number of pairs in the pool}}
	\State take the next two strings from the pool
        \For{j=1:K}
	       \State select a value in the first string uniformly at random, $v_1$, set $\mu=$ its position in the string
	       \State sample $u$ from $\mathcal{N}(\mu, \sigma^2)$ 
       		\If {$u<0$}
		\State set $u=0$ 
	       		\Else \If { $u>n$}
	       		        \State set $u=n$
                  \EndIf	 
                  \EndIf    
	   \State select the $u^{\text{th}}$ value in the second parent: $v_2$
           \State set $v'_1 = \alpha v_1 + (1-\alpha)v_2$
           \State set $v'_2 = (1-\alpha)v_1 + \alpha v_2$
	       \State replace the values of these bits: $v_1 = v'_1, \ v_2 = v'_2$
        \EndFor
\EndFor    
\end{algorithmic}
\caption{The $\beta$KBS Operator}
\label{fig:tab3}
\end{figure*}
One drawback of this algorithm is that the gene location chosen in the second parent can be outside the allowable range (less than 0 or greater than $n$). In this case we have chosen to replace any outlier values with either 0 or $n$. This is computationally simpler than redrawing another random number, but can bias the selection of genes towards these two values. We will address this problem in our future work.  
\section{Experimental Investigation}
\label{sec:experiments}
\subsection{Test Suite}
\label{sec:functions}
In this paper we only concern ourself with unconstrained multidimensional multimodal (except the Paraboloid function) nonlinear problems. The problem is formulated as:
\begin{equation*} 
\text{min} \ f(\mathbf{x}), \ f: \mathbb{R}^n \to \mathbb{R}
\end{equation*}
We use a range of test functions from the BBOB-2013 (see \cite{finck2010real}) list of noiseless functions. Most are well-known examples of minimization problems. Unless otherwise mentioned, the global minimum of the function is at $f(\mathbf{0}) = 0$. All bold values are vectors, i.e. $\mathbf{0} = (0,0, \ldots, 0)$, $\| \mathbf{x} \|$ is a square of a vector norm, $\sum_{k=1}^{n} x^2_k$.
In the experiments reported in the next section we use values of $n$ in the set $\{2, 5, 10, 20, 50\}$.
\begin{description}
\item[Paraboloid]
This is the only unimodal function we use 
\begin{equation*}
f_P(\mathbf{x})=\sum_{k=1}^{n} x^2_{k} = \| \mathbf{x} \|^2, \ -10 \leq x_k \leq 10
\end{equation*}
\item[Rastrigin] This function has $\approx 10^n$ local optima.
\begin{equation*}
f_R(\mathbf{x})= \| \mathbf{x} \|^2 + 10n - 10 \sum_{k=1}^{n}\cos (2 \pi k), \ -5.12 \leq x_k \leq 5.12 
\end{equation*}
\item[Rosenbrock] 
\begin{equation*}
f_{Ro}(\mathbf{x}) = \sum_{k=1}^{n-1}(100(x^2_{k+1}-x_k)^2 - (x_k-1)^2),\ -5.12 \leq x_k \leq 5.12
\end{equation*}
\noindent with the global minimum at $f(\mathbf{1})=0$ (where $\mathbf{1}$ denotes a vector consisting entirely of 1s). 
\item[Schwefel]
\begin{equation*}
f_S(\mathbf{x})=418.982 n - \sum_{k=1}^{n} x_k \sin \sqrt{|x_k|}, -500 \leq x_k \leq 500
\end{equation*}
with the global solution $f(\mathbf{\hat{x}}) = 0$ at $\mathbf{\hat{x}} = (420.9687, 420.9687, \ldots, 420.9687)$
\item[Ackley]
\begin{equation*}
f_A(\mathbf{x}) = 20 + e -20 \exp\left(-0.2 \sqrt{\frac{\| \mathbf{x} \|^2 }{n}}\right) - \exp \left(\frac{1}{n}\sum_{k=1}^{n}\cos(2 \pi x_k)\right),\ -32 \leq x_k \leq 32 
\end{equation*}
\item[Griewangk]
\begin{equation*}
f_G(\mathbf{x}) = 1 +\frac{1}{4000}\| \mathbf{x} \|^2 - \prod_{k=1}^{n} \cos \Bigg( \frac{x_k}{\sqrt{k}} \Bigg), \ -600 \leq x_k \leq 600
\end{equation*}
\end{description}
\subsection{$k$-Means Clustering problem}
The $k$-means clustering problem is a very well-known NP-hard problem that arises in many machine learning and signal processing applications. Given a set of $n$ vectors $\mathbf{x}_i \in \mathbb{R}^d (i=1, \ldots n)$, the problem is to find a partition of the data into $k$ clusters ($k < n$), each represented by the centroid of the data in that partition ($\mu_j \in \mathbb{R}^d, j=1, \ldots k$), the $k^{\text{th}}$ mean, such that the sum of some metric distances, typically the $L_2$ norm, between all data points $\mathbf{x}_i \forall i$ and the nearest centroid is minimized. In this paper we use Euclidean distance metric:
\begin{equation}
d_{X,Y} = \sum_{j=1}^{k} \sqrt{\sum_{i=1}^{s_j}  (x_i -  \mu_j)^2}
\end{equation}
Here we use $Y$ to denote the set of all centroids (means), $x_i$ and $\mu_j$ as defined above and $s_j$ is the number of vectors/observations in the $j^{\text{th}}$ subset. There are many local minima in the space, and the standard iterative algorithm (Lloyd's algorithm) typically finds a good one. There has been a lot of interest in heuristic algorithms solving this problem; some well-known implementations include \cite{kanungo2002efficient, hartigan1979algorithm, wagstaff2001constrained}. One of the best-known examples of $k$-means optimized by a GA is \cite{krishna1999genetic}. \\
\linebreak
In our implementation, we consider only the 4-means problem ($k=4$), and generate a set of random points $\mathbf{x}_i$ for each dimension and use the same data for all of the algorithms to allow fair comparison. We also report results of using Lloyd's algorithm using the implementation in the Scikit machine learning library for Python.
\subsection{Experimental Setup}
\label{sec:exsetup}
Each experiment consists of running every algorithm 20 times (from different randomized starting conditions) on a particular function with a pre-chosen dimensionality for 5000 generations. For simplicity, all algorithms save a single elite individual at each generation (i.e., the fittest individual is saved at every generation and added to the next generation of individuals, at the same time a randomly chosen offspring is deleted from the pool; thus population size is constant), and we do not vary the size of the population or the recombination pool, which were both fixed at 400 (making this a $(400+400)$EA). Tournament selection was used to select individuals for the recombination pool (see \cite{tersarkisov2012thesis} for its description). As well as KBS, we also used two variants of crossover that are well-known in the RCGA literature, BLX-$\alpha$ and SBX. For all recombination operators we use a $100 \%$ rate (recombine all pairs in the pool). We also applied two different mutation operators, uniform and Gaussian. In both cases bit mutation probability of $\frac{1}{n}$ was used, so that on the average there was one mutation per string. For the KBS both  $\alpha$ and $\beta$ were set to 0.4, and $K = \floor*{\frac{n}{2}}$. These values were chosen to make the changes introduced by the KBS operator comparable to those of other crossovers. All values are initiated uniformly at random in the interval specified in Section\ref{sec:functions}. 
In addition to BLX-$\alpha$ crossover explained in \ref{sec:theory}, we use Simulated Binary Crossover (SBX, see \cite{deb1995self, agrawal1994simulated}):
\begin{figure}

\begin{algorithmic}
	\State set $\eta$
	\State sample $u \sim U (0,1)$
	\If {$u<0.5$}
		\State set  $\beta  = (2 u)^{\frac{1}{\eta +1}}$
	\Else
		\State set  $\beta  = (2(1-u))^{-\frac{1}{\eta+1}}$
	\EndIf 
		\State $x'_k = 0.5 (x_k(1-\beta) + y_k (1+\beta))$
		\State $y'_k = 0.5 (x_k(1+\beta) + y_k (1-\beta))$
\end{algorithmic}
\caption{SBX operator}
\label{fig:sbx}
\end{figure}
For our experiments we set $\eta = 2$. We also use the following mutation operators in combination (to reduce bias) with crossovers:
\begin{enumerate}
\item Gaussian mutation (GM, see \cite{yoon2012roles}): in our implementation we select a bit $x_k$ with probability $\frac{1}{n}$ and sample from the normal distribution to generate the new value:
\begin{equation*}
x'_k = \frac{u_1 +u_2}{2} + \sqrt{u_2 - u_1} \cdot Z
\end{equation*}
where $Z \sim N(0,1), \ u_2$ is the maximal value of the bit and $u_1$ is the minimal value. 
\item Simple mutation (SM, each bit is sampled with probability $\frac{1}{n}$):
\begin{equation*} 
x'_k = (u_2 - u_1) \cdot U + u_1
\end{equation*}
where $U \sim U[0,1]$ is a uniform random variable. 
\end{enumerate}
We defined success as finding the global minimum to within a tolerance of $\varepsilon$. We set for 2-dimensional problems $\varepsilon = 0.01$, and $\varepsilon=0.1$ for all of the other problems. If the algorithm was successful, then the generation at which it first happened was recorded. The three measures that we record to establish the efficiency of the algorithm $A$ solving problem $\pi$ (where $j$ is $j^{\textnormal{th}}$ run out of the total of $R$ runs, $x^{\ast}$ is the best solution, and $t_j$ is the number of generations it took the algorithm to find the solution) are:
\begin{align}
I_{j} &= \Bigg\{ 
\begin{array}{cc}
1 & \textnormal{ if } f(x^{\ast})<\varepsilon,  \ \\
0 & \textnormal { otherwise } 
\end{array} \label{eq:measure3} \\
P(A,\pi) &= \frac{\sum_{j=1}^{R}I_j }{R}, \label{eq:measure1}\\
M(A, \pi) &= \frac{\sum_{j=1}^{R} t_j I_j}{R} \label{eq:measure2}
\end{align}
\subsection{Experimental Results}
The main results of this article are summarized in Tables \ref{tab:abs_res_threshold}, \ref{tab:abs_res_mean} and \ref{tab:abs_res_mean_kmeans}. In Table \ref{tab:abs_res_threshold} in every cell (crossover and mutation types vs function and dimension) the first value is the proportion of successful runs (Equation \ref{eq:measure1}). The second value is mean runtime averaged (applies to successful runs only), computed using Equation \ref{eq:measure2}. A run is considered successful if the $\varepsilon$-tolerance is reached. Tables \ref{tab:abs_res_mean} and \ref{tab:abs_res_mean_kmeans} simply display the mean fitness value at the end of the run, averaged over 20 runs.\\    
\linebreak
By looking at Tables \ref{tab:abs_res_threshold} and \ref{tab:abs_res_mean} we get a good general overview of the performance of algorithms on black-box optimization problems: we know which one does better on which function and can compare rate of convergence, since the faster the algorithm detects the basin of attraction around the global optimum, the better. Further analysis enables us to identify places where the operators tend to lead to convergence to local minima; for example both KBS operators appear to suffer from this (in 10 or more dimensions) for the Rosenbrock and Schwefel function. In contrast, both BLX-$\alpha$ and SBX suffer from slow convergence on these problems: it takes them much longer to explore the landscape segments with promising solutions.\\
\linebreak
It is clear from Tables \ref{tab:abs_res_threshold} - \ref{tab:abs_res_mean_kmeans} that most 2-dimensional problems were very simple, but in most standard optimization problems, KBS algorithms outperform the mainstream crossovers, possibly because they exploit current knowledge more effectively. They do especially well on 20- and 50-dimensional Ackley and Griewangk functions, which we see as a useful contribution to RCGA development if we compare the results, for example, to those in \cite{garcia2009study}.\\
\linebreak
By looking at the 4-means clustering problem (all dimensions), the results are reversed: KBS does worse compared to both BLX-$\alpha$ and SBX, quickly converging to a local solution and not being able to jump out of it. Although other operators also converge prematurely, this takes rather longer than for KBS, resulting in better solutions. As with the Schwefel function, we attribute this to the lack of exploration capacity of both KBS operators, something that needs to be fixed in the future work.\\
\linebreak
Nevertheless, all presented algorithms greatly outperform the k-means algorithm in SciKit module for Python, as is shown in Table \ref{tab:abs_res_mean_kmeans}.
One surprising result we encountered with Griewangk function (dimensions 2 and 5): Gaussian mutation improved all results by up to a factor of 20, something we did not observe with any other function in our experimentation. Although we attribute it to the structure of the fitness landscape, other factors may be at play that require additional investigation.  
\begin{landscape}
\begin{table*}
\hspace{-75pt}
\begin{tabular}{|c|c|c|c|c|c|c|c|c|c|c|c|c|c|c|c|c|c|}
\hline
\multirow{2}{*}{Function}&\multirow{2}{*}{Dims}&\multicolumn{4}{|c|}{$\alpha$KBS+}&\multicolumn{4}{|c|}{$\beta$KBS+} &\multicolumn{4}{|c|}{BLX-$\alpha$+}&\multicolumn{4}{|c|}{SBX+}\\
&&\multicolumn{2}{|c|}{SM}&\multicolumn{2}{|c|}{GM}&\multicolumn{2}{|c|}{SM}&\multicolumn{2}{|c|}{GM}&\multicolumn{2}{|c|}{SM}&\multicolumn{2}{|c|}{GM}&\multicolumn{2}{|c|}{SM}&\multicolumn{2}{|c|}{GM}\\
\hline
\multirow{5}{*}{Paraboloid}&2&1&7.2&1&4.75&1&6.95&1&4.5&1&13.7&1&6.35&1&13.65&1&5.75\\  
&5&1&44.8&1&17.2&1&45.3&1&18.5&1&1643.8&1&233.15&1&120.15&1&60.2\\
&10&1&134.65&1&53.7&1&100.15&1&46.75&0&-&0&-&0.8&2563&1&1250\\
&20&1&272.45&1&94.55&1&382.55&1&99.35&0&-&0&-&0&-&0&-\\
&50&1&746&1&240&1&866&1&233&0&-&0&-&0&-&0&-\\
\hline
\multirow{5}{*}{Rosenbrock}&2&1&20.5&1&25&1&32.4&1&27.45&1&46.1&1&38.6&1&35.45&1&31.15\\  
&5&0.05&2825&0&-&0.05&4546&0.05&1424&0&-&0&-&0.05&4816&0&-\\
&10&0&-&0&-&0&-&0&-&0&-&0&-&0&-&0&-\\
&20&0&-&0&-&0&-&0&-&0&-&0&-&0&-&0&-\\
&50&0&-&0&-&0&-&0&-&0&-&0&-&0&-&0&-\\
\hline
\multirow{5}{*}{Ackley}&2&0.95&2450&1&1050&1&1930&1&1335&1&3995&0.35&2155&0.15&1658&0.85&2180\\  
&5&1&1431&1&571&1&1307&1&583&0&-&0&-&0&-&0.25&3422\\
&10&0.6&3324&1&2381&1&3211&0.05&1565&0&-&0&-&0&-&0&-\\
&20&0.25&3849&0.9&2536&0.25&3981&0.75&2320&0&-&0&-&0&-&0&-\\
&50&0.25&3652&0.95&2174&1&3991&0.7&2846&0&-&0&-&0&-&0&-\\
\hline
\multirow{5}{*}{Rastrigin}&2&1&183&1&143&1&176&1&138&1&546&1&424&1&301&1&350\\  
&5&1&685&1&726&1&594&1&655&0&-&0&-&0&-&0&-\\
&10&0.45&3303&0.7&3111&0.6&2872&0.75&2538&0&-&0&-&0&-&0&-\\
&20&0.1&4064&0.05&4196&0&-&0.1&4255&0&-&0&-&0&-&0&-\\
&50&0&-&0&-&0&-&0&-&0&-&0&-&0&-&0&-\\
\hline
\multirow{5}{*}{Schwefel}&2&1&41&1&17&1&23&1&20&1&34&1&30&1&36&1&23\\  
&5&0.05&1879&0.05&1541&0.1&2144&0&-&0&-&0&-&0&-&0.05&3864\\
&10&0&-&0&-&0&-&0&-&0&-&0&-&0&-&0&-\\
&20&0&-&0&-&0&-&0&-&0&-&0&-&0&-&0&-\\
&50&0&-&0&-&0&-&0&-&0&-&0&-&0&-&0&-\\
\hline
\multirow{5}{*}{Griewangk}&2&1&1388&1&44.4&1&1496&1&38&1&2362&1&102&.8&1610&1&65\\  
&5&0.9&2312&1&127&0.95&1970&1&105&0&-&0.95&1445&0&-&1&1224\\
&10&0&-&0.85&1385&0&-&1&1009&0&-&0&-&0&-&0&-\\
&20&0.05&4361&0.8&1187&0&-&0.7&961&0&-&0&-&0&-&0&-\\
&50&0.05&4830&1&862&0&-&1&757&0&-&0&-&0&-&0&-\\
\hline
\end{tabular}
\caption{Success rate (proportion of 20 trials in which the global minimum was reached within tolerance $\varepsilon$) and the mean number of generations required to reach it (mean runtime)}
\label{tab:abs_res_threshold}
\end{table*}
\end{landscape}    
\begin{table*}
\hspace{-50pt}
\begin{tabular}{|c|c|c|c|c|c|c|c|c|c|}
\hline
\multirow{2}{*}{Function}&\multirow{2}{*}{Dims}&\multicolumn{2}{|c|}{$\alpha$KBS+}&\multicolumn{2}{|c|}{$\beta$KBS+}&\multicolumn{2}{|c|}{BLX-$\alpha$ +}& \multicolumn{2}{|c|}{SBX +} \\
&&SM&GM&SM&GM&SM&GM&SM&GM\\
\hline
\multirow{5}{*}{Paraboloid}&2&3.63e-7&1.37e-7&2.98e-7&1.83e-7&2.00e-5&5.23e-6&7.21e-6&2.10e-6\\  
&5&1.13e-5&1.91e-6&3.26e-6&8.49e-7&0.045&0.018&0.003&0.0016\\
&10&0.0008&0.00025&0.0004&0.00016&0.875&0.373&0.07&0.03\\
&20&0.0022&0.00109&0.0049&0.001&8.102&3.616&0.67&0.31\\
&50&0.0078&0.003&0.012&0.006&80.76&48.16&7.78&3.87\\
\hline
\multirow{5}{*}{Rosenbrock}&2&2.84e-5&1.50e-5&4.40e-5&3.20e-5&7.47e-5&5.66e-5&2.80e-5&2.54e-5\\  
&5&0.49&1.043&0.83&0.80&2.48&2.018&1.53&1.20\\
&10&8.32&8.32&8.03&8.06&35.99&31.30&14.43&9.62\\
&20&18.86&18.78&18.97&18.74&403.57&330.97&82.83&64.14\\
&50&48.91&48.79&48.97&48.86&7606.79&6353.38&572.33&481.14\\
\hline
\multirow{5}{*}{Ackley}&2&0.003&9.66e-4&0.003&0.002&0.05&0.01&0.02&0.006\\  
&5&0.008&0.003&0.007&0.003&2.59&0.90&0.72&0.17\\
&10&0.11&0.035&0.07&0.019&5.11&2.84&2.42&0.76\\
&20&0.164&0.046&0.18&0.06&8.30&4.46&3.91&2.39\\
&50&0.165&0.044&0.25&0.08&12.62&7.76&6.41&4.03\\
\hline
\multirow{5}{*}{Rastrigin}&2&1.73e-5&2.42e-5&7.42e-6&1.01e-5&1e-3&5.19e-4&5.15e-4&3.71e-4\\  
&5&2.99e-4&1.65e-4&1.05e-4&1.02e-4&2.67&2.22&1.32&0.92\\
&10&0.62&0.44&0.33&0.16&22.83&20.08&14.70&10.63\\
&20&2.72&2.32&3.38&2.56&89.97&83.62&60.24&59.30\\
&50&9.93&3.28&10.095&4.46&376.58&366.07&315.46&301.69\\
\hline
\multirow{5}{*}{Schwefel}&2&2.85e-5&3.03e-5&2.2e-5&2.8e-5&7.2e-5&5.87e-5&4.2e-5&2.5e-5\\  
&5&0.67&0.73&0.75&0.81&2.39&2.103&1.27&1.21\\
&10&8.33&8.18&8.13&8.02&37.59&30.4&10.15&9.71\\
&20&18.84&18.84&18.85&18.80&424.11&310.51&83.05&67.81\\
&50&49.13&48.82&49.22&48.80&7678&6104&553.90&501.33\\
\hline
\multirow{5}{*}{Griewangk}&2&0.003&1.35e-5&0.004&3.26e-6&0.01&2.16e-4&0.007&1e-4\\  
&5&0.05&0.005&0.051&0.009&0.62&0.07&0.37&0.04\\
&10&0.49&0.03&0.42&0.02&1.76&0.47&1.04&0.35\\
&20&0.57&0.05&0.74&0.02&8.27&1.05&1.63&0.72\\
&50&0.59&0.007&0.78&0.01&71.18&1.75&7.49&1.07\\
\hline
\end{tabular}
\caption{Mean value of the objective function (after 5000 generations), averaged over 20 trials.} 
\label{tab:abs_res_mean}
\end{table*}    
\begin{table*}
\hspace{-60pt}
\begin{tabular}{|c|c|c|c|c|c|c|c|c|c|c|}
\hline
\multirow{2}{*}{Function}&\multirow{2}{*}{Dims}&$\alpha$KBS&$\alpha$KBS&$\beta$KBS&$\beta$KBS&BLX-$\alpha$  & BLX-$\alpha$  & SBX  & SBX &SciKit \\
&&+SM&+GM&+SM&+GM&+SM&+GM&+SM&+GM&\\
\hline
\multirow{5}{*}{4-Means-Clustering}&2&10.54&10.44&10.35&10.51&8.63&8.63&8.62&8.62&10.31\\  
&5&37.05&37.24&36.97&37.002&31.51&31.54&31.37&31.40&64.07\\
&10&61.44&61.61&61.33&61.23&50.20&50.17&49.78&49.94&167.52\\
&20&97.29&97.14&96.96&97.15&80.58&80.55&83.62&83.28&369.13\\
&50&159.24&159.35&159.25&159.24&143.47&142.37&150.87&148.77&1063.29\\
\hline
\end{tabular}
\caption{Mean value of the objective function (after 5000 generations) for the 4-means clustering problem (compared to the 4-means classifier in the SciKit toolbox for Python, averaged over 20 trials.}
\label{tab:abs_res_mean_kmeans}
\end{table*}    
Having established the absolute performance of algorithms, we now rank them based on these results. As our data is not Gaussian-distributed or heteroscedastic, we have chosen to use non-parametric statistical tests, specifically the Mann-Whitney U statistic (also known as the one-sided nonparametric Wilcoxon rank-sum test), which is defined as:
\begin{align*}
U &= \min \{U_1, U_2 \} \\
U_1 &= s_1 s_2  + \frac{s_1 (s_1 +1)}{2} -R_1\\ 
U_2 &= s_1 s_2  + \frac{s_2 (s_2 +1)}{2} -R_2
\end{align*}
\noindent where $s_1, s_2$ are sample sizes (equal in our case) and $R_1, R_2$ are the sums of ranks in each sample. A sample size of 20 (which is our case) is enough to give a reliable estimate of the statistical significance of the difference of the means.
The test returns a z-statistic from a normal distribution and $p$-value that is compared to the significance level of $\alpha = 0.05$. The sign of the z-statistic shows which sample's mean is smaller: (-) for the first and (+) for the second.\\
\linebreak  
For reasons of space the particular values are not reported, but the main results we obtain from this analysis can be summarized as follows:
\begin{enumerate}
\item On the overwhelming majority of the Functional Optimization tasks, KBS-based algorithms outperform both BLX-$\alpha$ and SBX, and this difference is statistically significant (i.e., systematic). Among the exceptions are 2-dimensional Rosenbrock and Schwefel and 5-dimensional Griewangk function, where SBX+GM is equally efficient, because the U-statistic values are not significantly different from 0.
\item The 4-means clustering problem is best approximated (the global solution is, of course, unknown) by SBX-based algorithms (up to $n=10$) and BLX-$\alpha$ algorithms (for dimensionality $n >10$).
\item Gaussian mutation improves working on many instances (but never on the 4-means problem). This is true for each of the 4 types of crossover. It boosts performance especially well on the Griewangk test functions (all dimensions)
\item The two variants of the KBS recombination operator are almost equally good (most differences are not statistically significant). Out of 35 instances (7 functions $\times$ 5 dimensions) $\alpha$ KBS+SM outperforms $\beta$ KBS+SM in 4 instances, $\beta$ KBS+SM outperforms $\alpha$ KBS+SM in 5 instances, $\alpha$ KBS+GM outperforms $\beta$ KBS+GM in 6 instances and $\beta$ KBS+GM outperforms $\alpha KBS$+GM in 3 (the rest are not statistically significant). Overall, $\alpha KBS$ has a slight advantage over the other variant.   
\end{enumerate} 
Overall, we attribute the relative underperformance of BLX-$\alpha$ and SBX operators to our choice of elitism and selection function that prevent successful exploitation of promising fitness basins. We intend to address this problem in our future work.
\section{Conclusions and Future Work}
\label{sec:conclude}
In this article we have presented and tested a new recombination operator for RCGAs, a variant of the K-Bit-Swap that shares certain features with AX and BLX-$\alpha$ crossovers and Gaussian mutation. The principal difference is that in the KBS operator the locations of the bits selected in the two strings do not match, but are chosen randomly. We have considered two versions: chosen the two sites both uniformly at random, and using a normal distribution centered on the selected bit in the first string.\\ 
\linebreak
Both KBS operators have been shown to be superior for functional optimization problems to both BLX-$\alpha$ and SBX crossovers, but underperform on the 4-means approximation problem.\\
\linebreak
We also looked into some theoretical properties of presented operators. KBS samples different genes in both strings, thus slightly compensating for the absence of an exploration bias. If we consider uniform crossover and a simple 1-bit mutation (a standard choice for many applications), it is clear that fairly quickly the bits in the parents, $v_1$ and $v_2$ will be close, and even if we construct a certain interval around these values (as in BLX-$\alpha$ crossover), we can easily get stuck in some unpromising fitness region. KBS offers a workaround: although new values are not sampled from outside of the interval between the two parents, the second parent's value may be very different from the value in the same feature, thus it mimics exploration ability. Since the location selection for KBS is not restricted to the current feature, even if other features have already converged to the local optima, KBS has a relatively high probability of selecting a good schema and sampling in the area close to the optimal solution. Compared to the BLX-$\alpha$ operator new values lie strictly between the values selected in the parents. Therefore, both variants of the KBS operator are heavily biased towards exploitation rather than further exploration of the search space (see \cite{eshelman1993chapter,goldberg89genetic,mitchell96introduction}).\\ 
\linebreak
The logical next step would be to enhance the operator with an interval or other features that would enable generation of values outside of this interval. Also, to explore KBS properties further, we intend to study the selection pressure mechanisms that pushes evolution towards areas with high-quality schemata (in this article a simple tournament selection was used) and more sophisticated elitism functions (instead of the single fittest string). We believe that working along these lines will help improve performance of RCGAs on multidimensional and multimodal functions.
\bibliographystyle{abbrv}
\bibliography{kbsnew1}
\end{document}